\documentclass[10pt,twocolumn,letterpaper]{article}
\usepackage[pagenumbers]{cvpr}
\usepackage[T1]{fontenc}
\usepackage{amsmath,amssymb,booktabs,array,graphicx,multirow,tabularx}
\usepackage{xspace}
\usepackage{placeins}
\newcommand{\method}{SARVIA\xspace}

\DeclareMathOperator{\clip}{clip}
\DeclareMathOperator{\sgn}{sign}

\definecolor{cvprblue}{rgb}{0.21,0.49,0.74}
\usepackage[pagebackref,breaklinks,colorlinks,allcolors=cvprblue]{hyperref}
\def\confName{CVPR}
\def\confYear{2027}
\title{SARVIA: Fast-Adapting the Strong SAR Foundation Models That Are Already in Your MLLMs}
\author{
Yuxuan Li  \quad  Zichen Tian  \quad Qianru Sun$\dagger$ \\[3pt]
{\normalsize Singapore Management University} \\ 
{\normalsize  yuxuan.li.17@ucl.ac.uk; zichen.tian.2023@phdcs.smu.edu.sg; qianrusun@smu.edu.sg }
}
\begin{document}
\maketitle

\begin{abstract}
Synthetic aperture radar (SAR) foundation models provide transferable representations for diverse perception tasks in all-weather Earth observation.
However, their development often relies on costly large-scale SAR pretraining, raising the question of whether such extensive domain-specific pretraining is necessary. 
We find that modern multimodal large language models (MLLMs) already provide visual encoders that can outperform dedicated SAR foundation models as downstream initializations. 
This motivates a new paradigm: reuse general visual capabilities and devote domain training to brief specialization. 
We therefore introduce \method, an adaptive post-pretraining method that specializes modern MLLMs using self-supervised visual questions with computable answers. 
These questions are constructed from unlabeled observations through general image tasks that capture spatial relations and SAR-specific tasks that exploit native measurements and SAR--optical correspondence. 
A shared answer-prediction objective jointly trains low-rank adapters across visual and language components, with loss-driven task sampling. 
We then export the adapted visual encoder as a standalone backbone. 
With 600 adaptation steps and 2 GPU-hours, \method outperforms the compared foundation models on all six SAR detection and segmentation benchmarks. 
Compared with dedicated SAR pretraining, \method reduces domain-training costs by orders of magnitude while improving visual transfer across architectures and scales. 
This new paradigm shifts the emphasis from repeated, costly pretraining to a reusable specialization process that can benefit from advances in open MLLMs. 
Dataset and code: https://github.com/zcablii/SARVIA.
\end{abstract}

\section{Introduction}
\label{sec:intro}

\begin{figure}[!t]
\centering
\includegraphics[width=\linewidth,trim=29bp 36bp 28bp 23bp,clip]{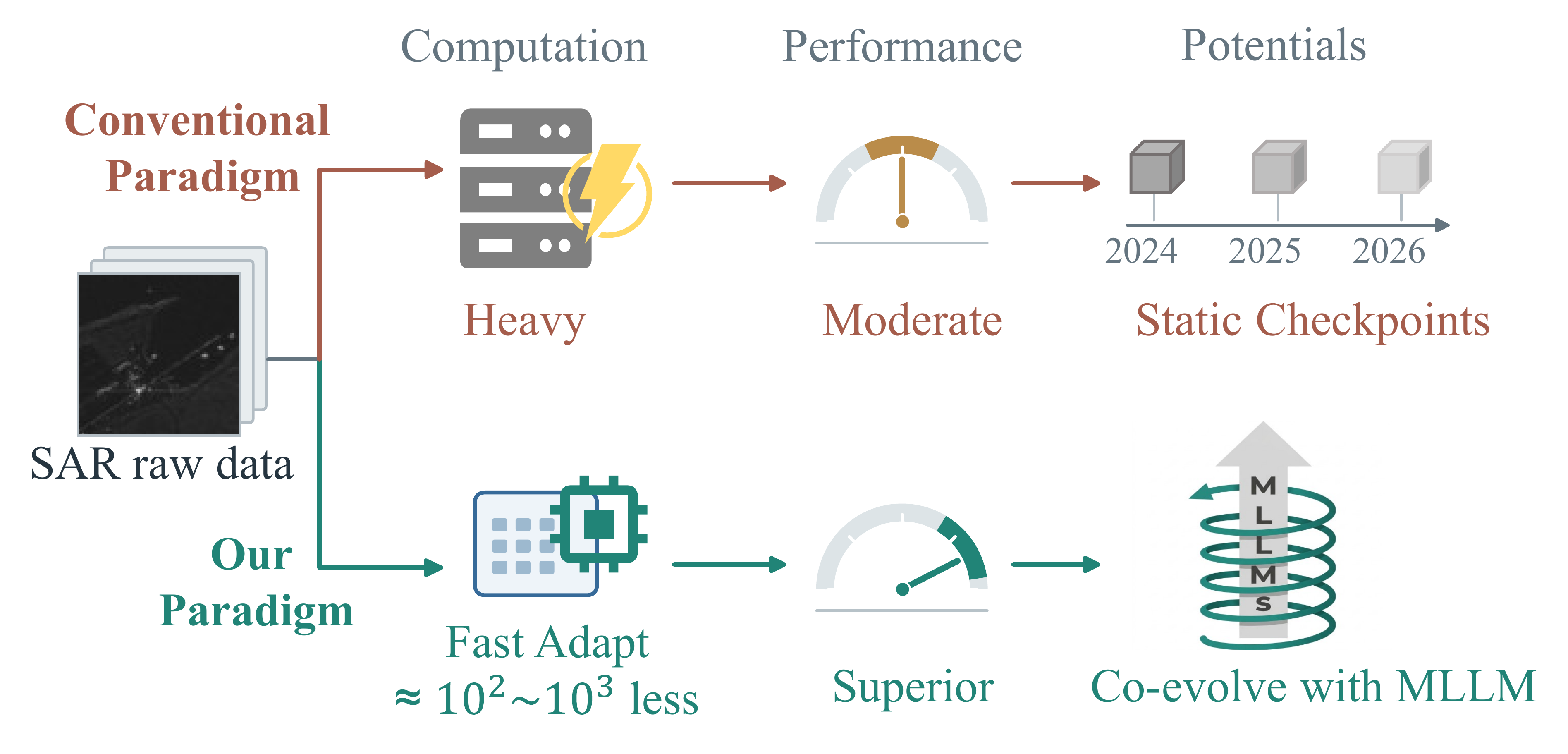}
\caption{\textbf{Paradigm shift from conventional dedicated pretraining to our proposed efficient adaptation.} \method improves SAR foundation models at lower pretraining cost and offers a reusable adaptation route as open MLLMs advance.}
\label{fig:paradigm}
\end{figure}

Synthetic aperture radar (SAR) foundation models provide shared visual representations for detection, segmentation, and other perception tasks in all-weather Earth observation~\cite{li2026saratrxv2,liu2026sarmae}. Learning these representations is challenging because SAR differs markedly from conventional photography: microwave scattering determines intensity, coherent interference produces speckle, and acquisition geometry changes object appearance~\cite{li2024sarjepa,li2026saratrxv2}. These properties have motivated radar-specific objectives and large-scale SAR pretraining~\cite{li2025saratrx,tang2026dinosar}. The resulting computational cost raises a central question: is extensive domain-specific pretraining still necessary when strong general visual encoders are already available?

Modern multimodal large language models (MLLMs) offer an alternative starting point. Their visual encoders benefit from broad visual learning and multimodal supervision~\cite{chen2024internvl25,bai2025qwen3vl,qwen2026qwen35}. We find that these encoders can already outperform dedicated foundation models as SAR downstream initializations. On our compact segmentation evaluation set, the original encoders extracted from Qwen3.5-2B and Qwen3.5-4B reach 36.29 and 38.89 mIoU, surpassing TerraMind~\cite{jakubik2025terramind} (34.69) and SARMAE~\cite{liu2026sarmae} (33.07) under the same protocol. Strong SAR features can therefore be inherited before additional domain pretraining.

This finding changes the role of SAR data from building a representation through extensive pretraining to specializing an already capable encoder. A short adaptation procedure can also be reused with successive open MLLMs, allowing SAR model development to benefit from advances in general visual learning (Fig.~\ref{fig:paradigm}). The practical challenge is then to adapt these encoders using unlabeled SAR observations, without constructing a manually annotated caption or question-answering corpus.

We address this challenge with \method, \emph{SAR Visual Adaptation}, an adaptive post-pretraining method that uses the MLLM's language component as a flexible task interface. Spatial indices, permutations, measurement ranks, and coordinates can all be expressed as text and learned through a shared answer-prediction objective. 
A new self-supervised task therefore requires a question generator and a computable answer, without a separate prediction head.

\method constructs two task families from unlabeled observations. 
General image tasks capture spatial relations through crop correspondence and patch ordering. 
SAR-specific tasks connect visible structure to native measurements, such as relative power and heterogeneity, and to corresponding regions in paired SAR and optical views. 
Known operations and measurements determine the answers. We jointly train LoRA adapters~\cite{hu2021lora} in the visual encoder, multimodal merger, and language model. 
To make effective use of the short adaptation budget, loss-driven sampling tracks each task's remaining loss relative to its initial level and recent learning progress. It allocates more examples to tasks with room to improve and evidence of learning, while reducing exposure to saturated or persistently stalled tasks. 
After adaptation, we merge the visual adapters and only export the encoder as a standalone backbone. 

\begin{figure}[!t]
\centering
\includegraphics[width=0.85\linewidth]{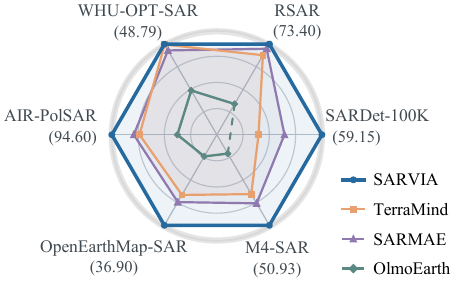}
\caption{\textbf{ViT-L downstream performance across six SAR benchmarks.} \method achieves the best score on every benchmark.}
\label{fig:downstream_radar}
\end{figure}

With 600 adaptation steps and approximately 2 RTX PRO 6000 GPU-hours, our ViT-L backbone leads the compared methods (Fig.~\ref{fig:downstream_radar}) on RSAR~\cite{zhang2025rsar}, SARDet-100K~\cite{li2024sardet100k}, M4-SAR~\cite{wang2026m4sar}, OpenEarthMap-SAR~\cite{xia2025openearthmapsar}, AIR-PolSAR-Seg-2.0~\cite{wang2025airpolsarseg2}, and WHU-OPT-SAR~\cite{li2022whuoptsar}. Domain-training cost is orders of magnitude lower than the dedicated pretraining baselines. The recipe improves every tested encoder across InternVL and Qwen, and stronger Qwen generations retain their advantage after adaptation. These results support a reusable specialization process that can advance with open MLLMs. Beyond SAR, the same framework could extend to other domains by replacing SAR-specific questions with computable domain-specific tasks while retaining the general image tasks and adaptation procedure. This offers a promising route to rapidly building specialized visual foundation models in domains where annotated instruction data are scarce. 

In summary, our contributions are:
\begin{itemize}
    \item We show that modern MLLM visual encoders already contain strong SAR capabilities, motivating a foundation-model paradigm based on general representation reuse and brief domain adaptation.
    \item We introduce \method, an efficient adaptive post-pretraining method that turns unlabeled observations into computable visual questions and exports standalone SAR visual backbones.
    \item We demonstrate \method shows leading ViT-L performance across six downstream benchmarks, with gains across MLLM families and scales at low training cost.
\end{itemize}

\section{Related Work}
\label{sec:related}

\begin{figure*}[!t]
\centering
\includegraphics[width=\textwidth,trim=8bp 7bp 14bp 10bp,clip]{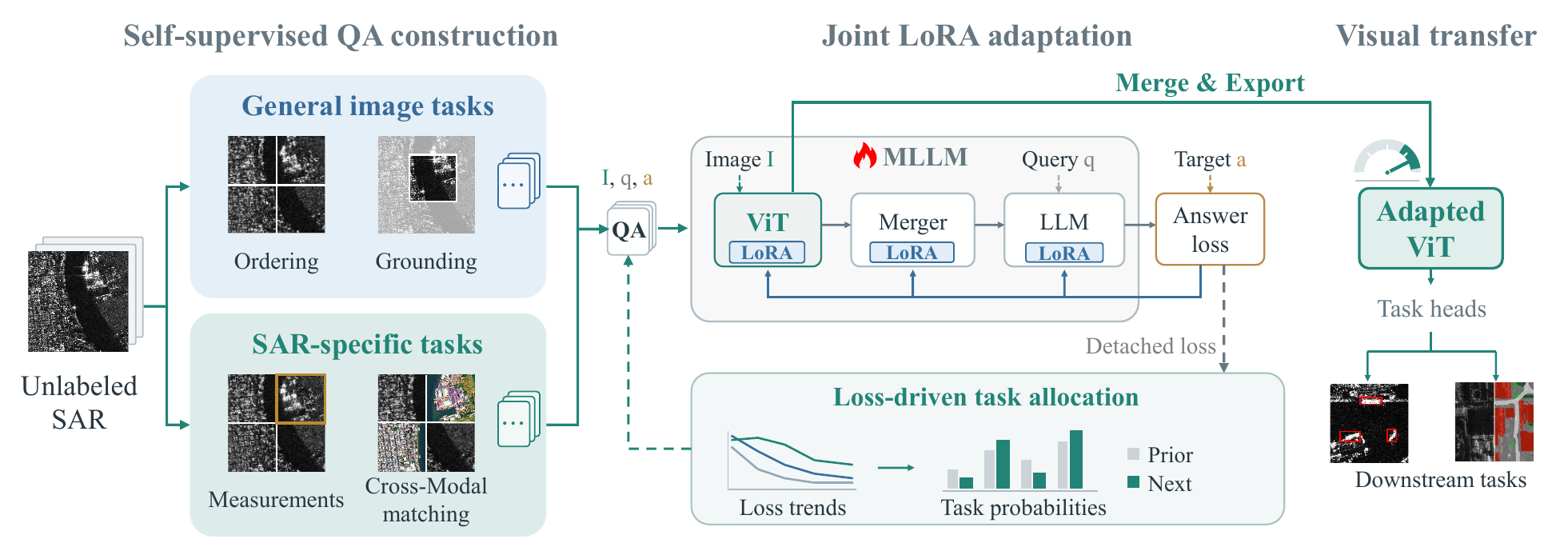}
\caption{\textbf{\method overview.} Unlabeled SAR observations yield tasks with deterministic answers. Answer loss jointly trains LoRA in the ViT, merger, and language model, while detached task losses adjust future QA draws through the loss-driven task allocation. We merge the visual adapters and export only the ViT for downstream perception.}
\label{fig:method}
\end{figure*}

\paragraph{SAR foundation models.}
SAR pretraining has developed along two complementary directions: expanding the diversity of training observations and designing objectives that reflect radar image formation. ATRNet-STAR broadens target-recognition coverage across scenes and acquisition conditions~\cite{liu2026atrnet}, while SARATR-X and DINOSAR combine larger SAR collections with domain-specific self-supervision~\cite{li2025saratrx,tang2026dinosar}. Objective design addresses the structure obscured by speckle: SAR-JEPA predicts gradient features, SARATR-X-v2 reconstructs structure across scales, and SUMMIT incorporates denoising and scattering-related tasks~\cite{li2024sarjepa,li2026saratrxv2,du2025summit}. SARMAE further combines speckle-aware augmentation with optical semantic anchors~\cite{liu2026sarmae}. Together, these efforts improve how representations are learned through large-scale dedicated SAR pretraining. Our work changes the starting point: we identify strong SAR features already present in open MLLMs and specialize them through brief adaptation.

\paragraph{Learning across Earth-observation modalities.}
Complementary sensors offer another source of supervision beyond the structure of a single SAR image. SatMAE exploits temporal and spectral information through masked reconstruction~\cite{cong2022satmae}. CROMA and TerraMind extend multimodal learning through radar--optical alignment and generative pretraining across geospatial modalities~\cite{fuller2023croma,jakubik2025terramind}. For detection, LSKNet models variable spatial context~\cite{li2025lsknet}, while multimodal approaches must also reconcile sensor-specific features and output formats. SM3Det addresses this through a shared mixture-of-experts model, whereas BabelRS uses language to align heterogeneous sensor features~\cite{li2026sm3det,pmlr-v306-li26q}. These approaches focus on learning representations that bring heterogeneous Earth-observation modalities together. \method instead turns their correspondence into supervision for specializing visual capabilities already acquired by general MLLMs, shifting the domain-training effort toward brief adaptation that can be reused as those models advance.

\paragraph{Adapting visual encoders through language.}
The shared image--text interface of MLLMs allows diverse visual tasks to supervise the same encoder~\cite{chen2024internvl25,bai2025qwen3vl,qwen2026qwen35,hopd}. ViTP applies this idea to domain-specific backbone pretraining: it trains a VLM through visual instructions and extracts the encoder for downstream perception~\cite{li2025vitp}. Its training mixture draws on existing captioning, VQA, grounding, and classification datasets, making the available annotated resources central to adaptation. \method retains the flexible language interface but generates both questions and answers from unlabeled observations, known transformations, and native measurements. It therefore removes the need for annotated instruction data during domain adaptation. This lowers the data requirements for extending the recipe to domains with scarce VQA resources, where suitable computable domain tasks could supply the supervision.

\paragraph{Computable supervision and task allocation.}
Spatial pretext tasks show how known image operations can replace manual labels: relative-position prediction and jigsaw solving derive their targets from patch locations and permutations~\cite{doersch2015context,noroozi2016jigsaw}. \method brings this principle into the MLLM interface, combining spatial questions with SAR measurements and cross-modal correspondence under one answer-prediction objective. A diverse task pool then raises an allocation problem: which questions should receive the limited adaptation budget? Curriculum methods use learning progress or deterioration to select tasks~\cite{matiisen2017tscl,graves2017acl}, while Dynamic Weight Average adjusts loss weights according to their rates of change~\cite{liu2019dwa}. Our sampler combines recent progress with normalized residual loss to distinguish saturation from persistent stagnation, reallocating examples within fixed source budgets rather than changing the shared prediction loss.

\section{SARVIA}
\label{sec:method}

\subsection{Self-supervised learning from computable answers}
\label{sec:formulation}

\method specializes an MLLM's visual encoder by asking questions whose answers can be computed from unlabeled observations (Fig.~\ref{fig:method}). The language interface lets spatial relations and SAR measurements share one training objective, so new tasks require neither manual VQA annotations nor separate prediction heads. We first construct this supervision, then adapt the MLLM and export its visual encoder for downstream perception.

Given a source record $z$, a task $\tau$, and construction parameters $\omega$, a deterministic generator produces $G_\tau(z;\omega)=(\mathcal I,q,a)$: an ordered image list, a question, and its answer. The parameters specify operations such as cropping or patch shuffling. Known geometry and measurements determine the answer, expressed as an index, permutation, rank, or bounding box. Varying the construction produces diverse questions while keeping each target computable from the source.

The MLLM learns to predict these answers by minimizing
\begin{equation}
 \mathcal L_{\rm ans}=-\sum_{k=1}^{|a|}
 \log P(a_k\mid\mathcal I,q,a_{<k}),
 \label{eq:answer_loss}
\end{equation}
where $a_k$ is the $k$-th answer token and $a_{<k}$ contains the preceding tokens. Only answer tokens contribute to the loss. This common objective accommodates two complementary task families: general image tasks teach spatial relationships, while SAR-specific tasks connect them to radar measurements and cross-sensor correspondence.

\subsection{General image tasks}
\label{sec:spatial_tasks}

Spatial relationships provide supervision across sensors without requiring semantic labels. Following spatial pretext learning~\cite{doersch2015context,noroozi2016jigsaw}, we ask the model to restore shuffled patches, match or locate a crop, select a missing region, or identify a flipped patch. For example, patch ordering requires recovering a known permutation by relating visible regions to a reference image. Such questions encourage the encoder to retain local structure across changes in position, arrangement, and scale.

These tasks operate within a modality and can be constructed from each new source. Varying crop scales and reference information changes the difficulty, while visibility and ambiguity checks exclude degenerate examples. Table~\ref{tab:supp_task_coverage} in the supplementary material lists the variants and their source coverage.

\subsection{SAR-specific tasks}
\label{sec:sar_tasks}

Spatial consistency alone does not capture the measurement properties of SAR. We therefore add questions that relate visible structure to native radar values and to corresponding optical observations.

\emph{Measurement tasks} ask the model to compare regions by power, heterogeneity, scattering concentration, edges, or spatial extent, with polarization and temporal comparisons where available. Relative comparisons expose radar-specific structure without requiring semantic scene labels. Native measurements define physical-property answers, while rendered images provide the visual input. Brightness and contrast questions instead use displayed pixel values. Keeping these quantities distinct ties each answer to the property named in the question.

\emph{SAR--optical tasks} require matching the same scene structure across different appearances. Using official SARLO pairs~\cite{debuysere2026sarlo}, we construct bidirectional crop matching, ordering, grounding, missing-region selection, and synthetic-shift correction tasks. The official pair grid supplies the geometric targets. Compositional questions combine these relations, for example by locating a region across modalities before comparing its properties. These questions complement the general image tasks by grounding spatial relationships in SAR-specific evidence.

\subsection{Adaptation and visual transfer}
\label{sec:adaptation}

We jointly train LoRA adapters~\cite{hu2021lora} in the visual encoder, multimodal merger, and language model, keeping the base weights frozen. Updating the visual encoder specializes its features, while adapting the merger and decoder lets the question--answer interface accommodate SAR inputs and task formats. All tasks share the adapters and the objective in Eq.~\ref{eq:answer_loss}.

\paragraph{Loss-driven task allocation.}
The tasks learn at different rates: some quickly saturate, others improve steadily, and some remain near their initial loss. With a short adaptation budget, high loss alone is therefore an insufficient reason to allocate more examples. We favor tasks with both room to improve and evidence of learning, while restoring attention when performance deteriorates.

For each source--task variant $g$, let $F_g$ and $S_g$ denote fast and slow exponential moving averages of its mean answer-token loss, and $R_g$ its initial loss. The normalized residual $r_g=\min(1,F_g/R_g)$ measures how much of the initial loss remains. A progress score $q_g\in[-1,1]$, computed from the noise-thresholded relative gap between $S_g$ and $F_g$, is positive when loss falls and negative when it rebounds. We combine the two signals into
\begin{equation}
 u_g=4r_g(1-r_g)+\lambda_p[q_g]_+r_g
             +\lambda_f[-q_g]_+,
 \label{eq:utility}
\end{equation}
where $[x]_+=\max(x,0)$ and $\lambda_p,\lambda_f$ weight progress and rebound. The first term favors intermediate residuals over saturation or persistent stagnation. The second rewards ongoing improvement. The third revisits tasks whose loss rises.

We redistribute draws within each source to preserve the sensor mixture. Let $p_g^0$ be the initial probability of variant $g$, $\mathcal G_s$ the variants from source $s$, and $M_s=\sum_{h\in\mathcal G_s}p_h^0$ that source's budget. The target probability is
\begin{equation}
 d_g=\eta p_g^0+(1-\eta)M_s
 \frac{p_g^0u_g}{\sum_{h\in\mathcal G_s}p_h^0u_h},
 \label{eq:sampling}
\end{equation}
where $\eta\in[0,1]$ retains a fraction of the initial mixture for exploration. The remaining draws follow prior-weighted utility. Feedback comes from detached training losses, requiring no extra model evaluations. The supplementary material gives the moving-average updates and probability stabilization details.

\begin{table*}[!ht]
\centering
\small
\setlength{\tabcolsep}{3pt}
\begin{tabular}{@{}llccrrrlcc@{}}
\toprule
Dataset & Task & Categories & Metric & Train & Val. & Test & Prediction head & Input / crop & Schedule \\
\midrule
RSAR & Oriented det. & 6 & AP$_{50}$ & 78,837 & -- & 8,538 & Oriented R-CNN~\cite{xie2021orientedrcnn} & $800^2$ & 12 epochs \\
M4-SAR & Oriented det. & 6 & AP$_{50}$ & 56,116 & 22,112 & 33,946 & Oriented R-CNN~\cite{xie2021orientedrcnn} & $512^2$ & 12 epochs \\
SARDet-100K & Horizontal det. & 6 & mAP & 94,493 & 10,492 & 11,613 & Cascade R-CNN~\cite{cai2018cascadercnn} & $1024^2$ & 12 epochs \\
OpenEarthMap-SAR & Segmentation & 8 & mIoU & 4,333 & 210 & 490 & UPerNet~\cite{xiao2018upernet} & $512^2$ & 40k steps \\
AIR-PolSAR-Seg-2.0 & Segmentation & 6 & mIoU & 1,233 & 309 & -- & UPerNet~\cite{xiao2018upernet} & $512^2$ & 40k steps \\
WHU-OPT-SAR & Segmentation & 7 & mIoU & 17,640 & 5,880 & 5,880 & UPerNet~\cite{xiao2018upernet} & $256^2$ & 40k steps \\
\bottomrule
\end{tabular}
\caption{Downstream benchmark settings. Evaluation uses the listed test partitions, except AIR-PolSAR-Seg-2.0 (validation). }
\label{tab:downstream_setup}
\end{table*}

\paragraph{Exporting the visual backbone.}
After adaptation, we merge the visual LoRA weights and extract the encoder before the multimodal merger. MLLM pretraining optimizes this encoder to supply visual tokens to a language model. Its single-scale token interface does not explicitly provide the local spatial priors or feature hierarchy needed for dense prediction. Following ViTP~\cite{li2025vitp}, we attach ViT-Adapter~\cite{chen2023vitadapter} to introduce local spatial inductive biases and construct multiscale features. The exported encoder, spatial adapter, and task head are then fine-tuned for detection or segmentation. The language model and multimodal merger are needed only during adaptation.
\label{sec:method_end}

\section{Experiments}
\label{sec:experiments}
\setlength{\textfloatsep}{12pt plus 2pt minus 2pt}

\subsection{Experimental setup}
\label{sec:experimental_setup}

\paragraph{Pretraining data and adaptation.}
We construct 87 task variants from the collected TerraMesh~\cite{blumenstiel2025terramesh}, SARLO-80~\cite{debuysere2026sarlo}, and KuroSiwo~\cite{bountos2024kurosiwo} datasets, covering Sentinel-1 observations, high-resolution SAR--optical pairs, and multitemporal SAR. With the raw metadata, native measurements supply targets for domain-specific questions. We sample the three sources at a $1{:}4{:}1$ ratio, reflecting their relative task coverage, and adjust task probabilities within each source through loss-driven allocation.

Our default ViT-L model adapts Qwen3.5-2B~\cite{qwen2026qwen35} visual encoder with rank-16 LoRA and scaling parameter $\alpha=32$. Training uses 600 updates at a global batch size of 128, totaling 76,800 question--answer draws. ViT-B adapts Qwen3.5-0.8B visual encoder with rank-8 LoRA for the same number of updates. Table~\ref{tab:supp_training} in the supplementary material provides the full settings.

\paragraph{Development set and encoder comparisons.}
We use our established OEM-SLC-Amplitude for rapid encoder comparisons and ablations (Table~\ref{tab:oem_encoders}, Table~\ref{tab:sampling_comparison}, Table~\ref{tab:sampling_allocation}, Figure~\ref{fig:step_ablation} and Figure~\ref{fig:sampling_analysis}), then evaluate transfer on six public benchmarks. This development set contains 274 training and 654 evaluation chips constructed from native Umbra SLC amplitudes and projected OpenEarthMap-SAR labels~\cite{xia2025openearthmapsar}. All encoder comparisons fine-tune the standalone ViT with ViT-Adapter and UPerNet~\cite{chen2023vitadapter,xiao2018upernet} for 50 epochs under the same protocol, reporting the final fine-tuning checkpoint for every model. Section~\ref{sec:supp_oem} of the supplementary material describes the dataset construction.

\paragraph{Downstream benchmarks.}
The detection benchmarks test localization across object geometries and sensor conditions. \textbf{RSAR}~\cite{zhang2025rsar} provides oriented boxes for six categories with varied sizes and orientations. \textbf{M4-SAR}~\cite{wang2026m4sar} pairs Sentinel-1 VV/VH with optical imagery and annotates six infrastructure categories, including bridges, oil tanks, and airports. Our SAR-only evaluation tests extended structures across coastal and urban scenes. Both use rotated AP$_{50}$. \textbf{SARDet-100K}~\cite{li2024sardet100k} unifies ten datasets with resolutions of 0.1--25\,m and multiple polarization modes, testing six-category horizontal detection across heterogeneous sensors and scales with COCO mAP.
The segmentation benchmarks test dense prediction at different resolutions and polarization settings, using mIoU throughout. \textbf{OpenEarthMap-SAR}~\cite{xia2025openearthmapsar} provides eight-class labels for 0.15--0.5\,m Umbra imagery across 35 regions. Its VV or HH amplitudes test fine delineation of buildings, roads, and vegetation. \textbf{AIR-PolSAR-Seg-2.0}~\cite{wang2025airpolsarseg2} provides six-class terrain labels for 8\,m quad-polarimetric Gaofen-3 imagery of three Chinese regions. We use HH, HV, and VV amplitudes to assess polarization-dependent cues and irregular land-cover boundaries. \textbf{WHU-OPT-SAR}~\cite{li2022whuoptsar} contains seven-class labels for paired Gaofen-1/Gaofen-3 imagery at 5\,m resolution in Hubei. Its SAR inputs test contextual distinctions such as city versus village, alongside natural land cover and narrow roads.

All downstream tasks fine-tune the exported ViT, ViT-Adapter, and task head using the settings in Table~\ref{tab:downstream_setup}. For these 6 downstream benchmarks, we report the performance of the final checkpoint. 

\subsection{Strong initializations and efficient adaptation}
\label{sec:encoder_comparison}

We first separate the capability inherited from an MLLM from the gain provided by SAR adaptation (Table~\ref{tab:oem_encoders}). The unadapted Qwen3.5-2B and Qwen3.5-4B encoders already exceed TerraMind and SARMAE under the same downstream protocol. Thus, extensive SAR pretraining is not a prerequisite for obtaining a strong SAR initialization: general multimodal learning can supply one before any additional domain training. Results from the official Qwen3.5-2B encoder on all six public benchmarks further support this finding (Table~\ref{tab:supp_vanilla_transfer} in the supplementary material).

\paragraph{Generality across architectures and scales.}
\method consistently improves all five encoders across InternVL2.5, Qwen3-VL, and Qwen3.5, whose parent MLLMs span 0.8B--4B parameters. Notably, the gains are not limited to weaker starting points: even the strongest original encoder benefits from our approach. These results demonstrate that our method is effective and exhibits strong generalization ability across different MLLM families and parameter scales.

\begin{table}[!t]
\centering
\small
\setlength{\tabcolsep}{1.5pt}
\begin{tabular}{@{}lcccc@{}}
\toprule
Encoder source & Backbone & Vanilla & +\method & GPU-hours \\
\midrule
\multicolumn{5}{l}{\textit{SAR / Earth-observation foundation models}} \\
TerraMind~\cite{jakubik2025terramind} & ViT-L & 34.69 & \textemdash & $\approx$5,122 \\
SARMAE~\cite{liu2026sarmae} & ViT-L & 33.07 & \textemdash & $\approx$320 \\
\midrule
\multicolumn{5}{l}{\textit{MLLM visual encoders}} \\
Qwen3-VL-2B~\cite{bai2025qwen3vl} & ViT-L & 28.42 & 32.66 {\scriptsize(+4.24)} & 1.87 \\
InternVL2.5-2B~\cite{chen2024internvl25} & ViT-L & 33.79 & 35.37 {\scriptsize(+1.58)} & 2.71 \\
Qwen3.5-0.8B~\cite{qwen2026qwen35} & ViT-B & 33.30 & 35.54 {\scriptsize(+2.24)} & 0.95 \\
Qwen3.5-2B~\cite{qwen2026qwen35} & ViT-L & 36.29 & 39.70 {\scriptsize(+3.41)} & 1.92 \\
Qwen3.5-4B~\cite{qwen2026qwen35} & ViT-L & 38.89 & 39.95 {\scriptsize(+1.06)} & 2.64 \\
\bottomrule
\end{tabular}
\caption{Standalone encoder transfer on OEM-SLC-Amplitude (mIoU, \%). GPU-hours of pretraining are estimated or measured on a single RTX PRO 6000.}
\label{tab:oem_encoders}
\end{table}

\paragraph{Inheriting progress in general MLLMs.}
At the 2B scale, replacing Qwen3-VL with Qwen3.5 improves the original encoder by 7.87 mIoU and the adapted encoder by 7.04. Most of the advantage of the newer starting point survives specialization. This supports a practical upgrade strategy: improve the general initialization and reuse the SAR adaptation recipe, allowing progress in open MLLMs to carry into the domain backbone.

\begin{table}[!t]
\centering
\small
\setlength{\tabcolsep}{3pt}
\renewcommand{\arraystretch}{1.04}
\begin{tabular}{@{}>{\raggedright\arraybackslash}p{0.45\linewidth}>{\raggedright\arraybackslash}p{0.32\linewidth}c@{}}
\toprule
Method & Backbone & AP$_{50}$ $\uparrow$ \\
\midrule
O-RCNN~\cite{xie2021orientedrcnn} & ResNet-50 & 64.82 \\
SARMAE~\cite{liu2026sarmae} & ViT-B & 66.80 \\
RoI Transformer~\cite{ding2019roitransformer} & ResNet-50 & 66.95 \\
GSDet~\cite{ding2025gsdet} & ResNet-50 & 68.00 \\
\textbf{\method (Ours)} & ViT-B & 70.29 \\
\midrule\midrule
OlmoEarth~\cite{herzog2026olmoearth} & ViT-L & 57.93 \\
SatMAE~\cite{cong2022satmae} & ViT-L & 67.99 \\
RemoteCLIP~\cite{liu2024remoteclip} & ViT-L & 69.18 \\
TerraMind~\cite{jakubik2025terramind} & ViT-L & 70.53 \\
SARMAE~\cite{liu2026sarmae} & ViT-L & 72.20 \\
\textbf{\method (Ours)} & ViT-L & \textbf{73.40} \\
\bottomrule
\end{tabular}
\caption{Results of RSAR~\cite{zhang2025rsar} oriented detection.}
\label{tab:rsar_results}
\end{table}

\begin{table}[!t]
\centering
\small
\setlength{\tabcolsep}{3pt}
\renewcommand{\arraystretch}{1.04}
\begin{tabular}{@{}>{\raggedright\arraybackslash}p{0.45\linewidth}>{\raggedright\arraybackslash}p{0.32\linewidth}c@{}}
\toprule
Method & Backbone & mAP $\uparrow$ \\
\midrule
SUMMIT~\cite{du2025summit} & ViT-B & 50.86 \\
SARMAE~\cite{liu2026sarmae} & ViT-B & 52.00 \\
SARATR-X~\cite{li2025saratrx} & HiViT-B & 54.20 \\
DINO (UniSAR-7M)~\cite{caron2021dino,tang2026dinosar} & ViT-B & 54.22 \\
\textbf{\method (Ours)} & ViT-B & 55.10 \\
\midrule\midrule
SatMAE~\cite{cong2022satmae} & ViT-L & 40.98 \\
OlmoEarth~\cite{herzog2026olmoearth} & ViT-L & 42.11 \\
TerraMind~\cite{jakubik2025terramind} & ViT-L & 50.62 \\
SARMAE~\cite{liu2026sarmae} & ViT-L & 54.11 \\
\textbf{\method (Ours)} & ViT-L & \textbf{59.15} \\
\bottomrule
\end{tabular}
\caption{Results of SARDet-100K~\cite{li2024sardet100k} horizontal detection.}
\label{tab:sardet_results}
\end{table}
 
\begin{table}[!t]
\centering
\small
\setlength{\tabcolsep}{3pt}
\renewcommand{\arraystretch}{1.04}
\begin{tabular}{@{}>{\raggedright\arraybackslash}p{0.45\linewidth}>{\raggedright\arraybackslash}p{0.32\linewidth}c@{}}
\toprule
Method & Backbone & AP$_{50}$ $\uparrow$ \\
\midrule
OlmoEarth~\cite{herzog2026olmoearth} & ViT-B & 21.86 \\
TerraMind~\cite{jakubik2025terramind} & ViT-B & 39.12 \\
SUMMIT~\cite{du2025summit} & ViT-B & 40.18 \\
SARMAE~\cite{liu2026sarmae} & ViT-B & 40.73 \\
\textbf{\method (Ours)} & ViT-B & 41.23 \\
\midrule\midrule
SatMAE~\cite{cong2022satmae} & ViT-L & 23.26 \\
OlmoEarth~\cite{herzog2026olmoearth} & ViT-L & 26.45 \\
TerraMind~\cite{jakubik2025terramind} & ViT-L & 40.28 \\
SARMAE~\cite{liu2026sarmae} & ViT-L & 43.43 \\
SARATR-X-v2~\cite{li2026saratrxv2} & iTPN-L & {46.62} \\
\textbf{\method (Ours)} & ViT-L & \textbf{50.93} \\
\bottomrule
\end{tabular}
\caption{Results of M4-SAR~\cite{wang2026m4sar} oriented detection.}
\label{tab:m4sar_results}
\end{table}

\begin{table}[!t]
\centering
\small
\setlength{\tabcolsep}{3pt}
\renewcommand{\arraystretch}{1.04}
\begin{tabular}{@{}>{\raggedright\arraybackslash}p{0.45\linewidth}>{\raggedright\arraybackslash}p{0.32\linewidth}c@{}}
\toprule
Method & Backbone & mIoU $\uparrow$ \\
\midrule
SARATR-X-v2~\cite{li2026saratrxv2} & iTPN-B & 31.34 \\
VMamba~\cite{liu2024vmamba} & VSS & 34.74 \\
U-Net~\cite{ronneberger2015unet} & EfficientNet-B4 & 35.13 \\
\textbf{\method (Ours)} & ViT-B & 35.76 \\
\midrule\midrule
OlmoEarth~\cite{herzog2026olmoearth} & ViT-L & 27.85 \\
SatMAE~\cite{cong2022satmae} & ViT-L & 27.91 \\
TerraMind~\cite{jakubik2025terramind} & ViT-L & 33.51 \\
SARMAE~\cite{liu2026sarmae} & ViT-L & 32.85 \\
SARATR-X-v2~\cite{li2026saratrxv2} & iTPN-L & 35.14 \\
\textbf{\method (Ours)} & ViT-L & \textbf{36.90} \\
\bottomrule
\end{tabular}
\caption{Results of OpenEarthMap-SAR~\cite{xia2025openearthmapsar} segmentation.}
\label{tab:earthmap_results}
\end{table}

\begin{table}[!t]
\centering
\small
\setlength{\tabcolsep}{3pt}
\renewcommand{\arraystretch}{1.04}
\begin{tabular}{@{}>{\raggedright\arraybackslash}p{0.45\linewidth}>{\raggedright\arraybackslash}p{0.32\linewidth}c@{}}
\toprule
Method & Backbone & mIoU $\uparrow$ \\
\midrule
SARATR-X-v2~\cite{li2026saratrxv2} & iTPN-B & 83.43 \\
TerraSegNet~\cite{wijaya2025terrasegnet} & EfficientNetV2-S & {92.10} \\
\textbf{\method (Ours)} & ViT-B & 93.37 \\
\midrule\midrule
SatMAE~\cite{cong2022satmae} & ViT-L & 78.75 \\
OlmoEarth~\cite{herzog2026olmoearth} & ViT-L & 79.25 \\
TerraMind~\cite{jakubik2025terramind} & ViT-L & 88.12 \\
SARMAE~\cite{liu2026sarmae} & ViT-L & 89.46 \\
SARATR-X-v2~\cite{li2026saratrxv2} & iTPN-L & 90.20 \\
\textbf{\method (Ours)} & ViT-L & \textbf{94.60} \\
\bottomrule
\end{tabular}
\caption{Results of AIR-PolSAR-Seg-2.0~\cite{wang2025airpolsarseg2} segmentation.}
\label{tab:airpol_results}
\end{table}

\begin{table}[!t]
\centering
\small
\setlength{\tabcolsep}{3pt}
\renewcommand{\arraystretch}{1.04}
\begin{tabular}{@{}>{\raggedright\arraybackslash}p{0.45\linewidth}>{\raggedright\arraybackslash}p{0.32\linewidth}c@{}}
\toprule
Method & Backbone & mIoU $\uparrow$ \\
\midrule
PIDNet~\cite{xu2023pidnet} & PIDNet-S & 44.30 \\
OCRNet~\cite{yuan2020ocrnet} & HRNetV2-W48 & 44.80 \\
SegFormer~\cite{xie2021segformer} & MiT-B4 & 45.30 \\
SARATR-X-v2~\cite{li2026saratrxv2} & iTPN-B & 46.10 \\
Mask2Former~\cite{cheng2022mask2former} & ResNet-101 & 46.80 \\
TransUNet~\cite{chen2021transunet} & ResNet-50 + ViT & 47.10 \\
\textbf{\method (Ours)} & ViT-B & 48.19 \\
\midrule\midrule
OlmoEarth~\cite{herzog2026olmoearth} & ViT-L & 41.73 \\
SETR~\cite{zheng2021setr} & ViT-L & 44.50 \\
SatMAE~\cite{cong2022satmae} & ViT-L & 45.34 \\
SARATR-X-v2~\cite{li2026saratrxv2} & iTPN-L & 47.10 \\
SARMAE~\cite{liu2026sarmae} & ViT-L & 47.83 \\
TerraMind~\cite{jakubik2025terramind} & ViT-L & {48.72} \\
\textbf{\method (Ours)} & ViT-L & \textbf{48.79} \\
\bottomrule
\end{tabular}
\caption{Results of WHU-OPT-SAR~\cite{li2022whuoptsar} segmentation.}
\label{tab:whu_results}
\end{table}

\paragraph{Efficient SAR adaptation.}
Our main model requires just 600 adaptation steps and 1.92 RTX PRO 6000 GPU-hours. This is approximately $1/170$ of SARMAE's domain-training cost and $1/2{,}700$ of TerraMind's, while improving downstream mIoU by 6.63 and 5.01 points, respectively (Table~\ref{tab:oem_encoders}). The low cost therefore accompanies stronger transfer, rather than a sacrifice in performance. By inheriting general visual capabilities, \method concentrates the additional computation on SAR specialization, making high-performing domain backbones accessible with a small training budget.

\subsection{Performance on downstream SAR benchmarks}
\label{sec:main_results}
\suppressfloats[t]

We next test whether the advantage extends beyond the development setting. Tables~\ref{tab:rsar_results}--\ref{tab:whu_results} compare \method with other state-of-the-art SAR foundation models or Earth-observation foundation models. Lower blocks group ViT-L-scale backbones. Bold marks the best score.

\paragraph{Transfer across prediction formats.}
ViT-L leads all six comparisons, spanning oriented detection, horizontal detection, and segmentation. Its largest margins occur on M4-SAR and SARDet-100K, even though method development uses segmentation, supporting transfer beyond the development task. This advantage is not confined to the larger backbone: ViT-B exceeds every listed ViT-L foundation baseline on SARDet-100K, OpenEarthMap-SAR, and AIR-PolSAR. Brief specialization can therefore make a smaller inherited encoder more effective than larger dedicated models, strengthening the case for representation reuse over backbone size alone.

\subsection{Adaptation steps and LoRA rank}
\label{sec:step_ablation}

\begin{figure}[!t]
\centering
\includegraphics[width=\linewidth]{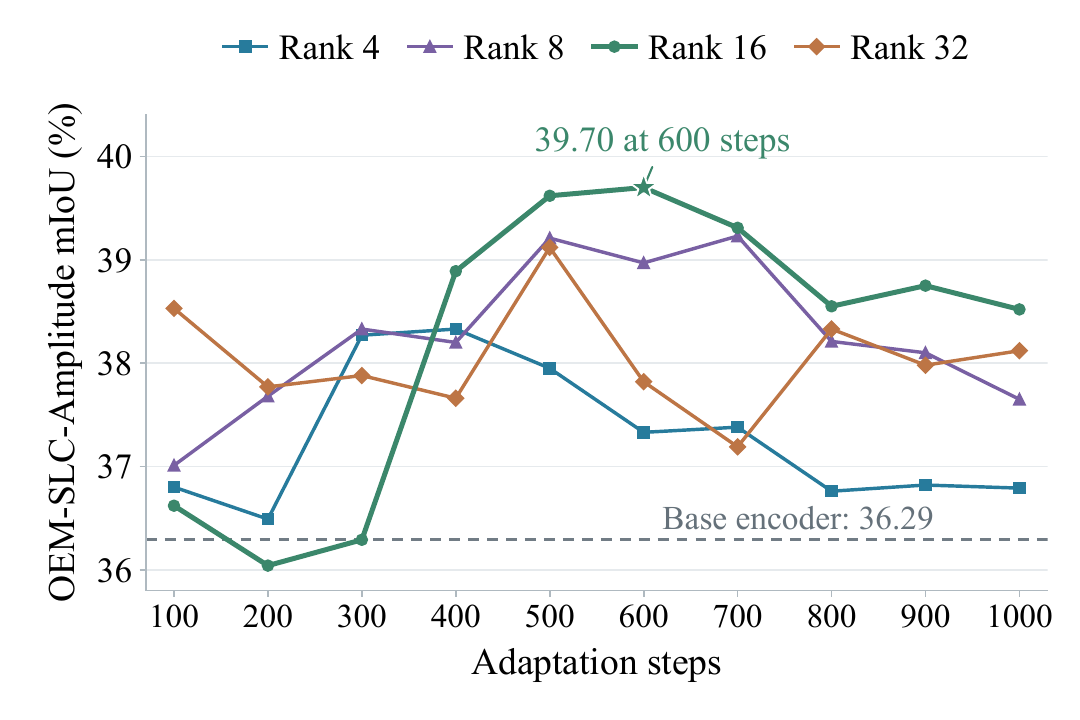}
\caption{\textbf{Adaptation length and rank on Qwen3.5-2B.} OEM-SLC-Amplitude mIoU for four LoRA ranks. The dashed line marks the original visual encoder.}
\label{fig:step_ablation}
\end{figure}

Figure~\ref{fig:step_ablation} varies adaptation length and LoRA rank on Qwen3.5-2B with $\alpha/r=2$. All four ranks reach their best transfer within 400--700 steps, consistent with specialization of an already capable encoder. Early progress does not predict the best final representation: rank 32 leads at 100 steps, but rank 16 peaks higher at step 600. Continuing rank-16 training to 1000 steps loses 1.18 mIoU, and doubling the rank lowers the best score by 0.58. More training or adapter capacity need not improve transfer. Qwen3.5-0.8B similarly reaches its best tested setting at step 600 with rank 8, which we use for ViT-B.

\subsection{Task families and sampling strategies}
\label{sec:sampling_comparison}

\begin{figure*}[t]
\centering
\includegraphics[width=\textwidth]{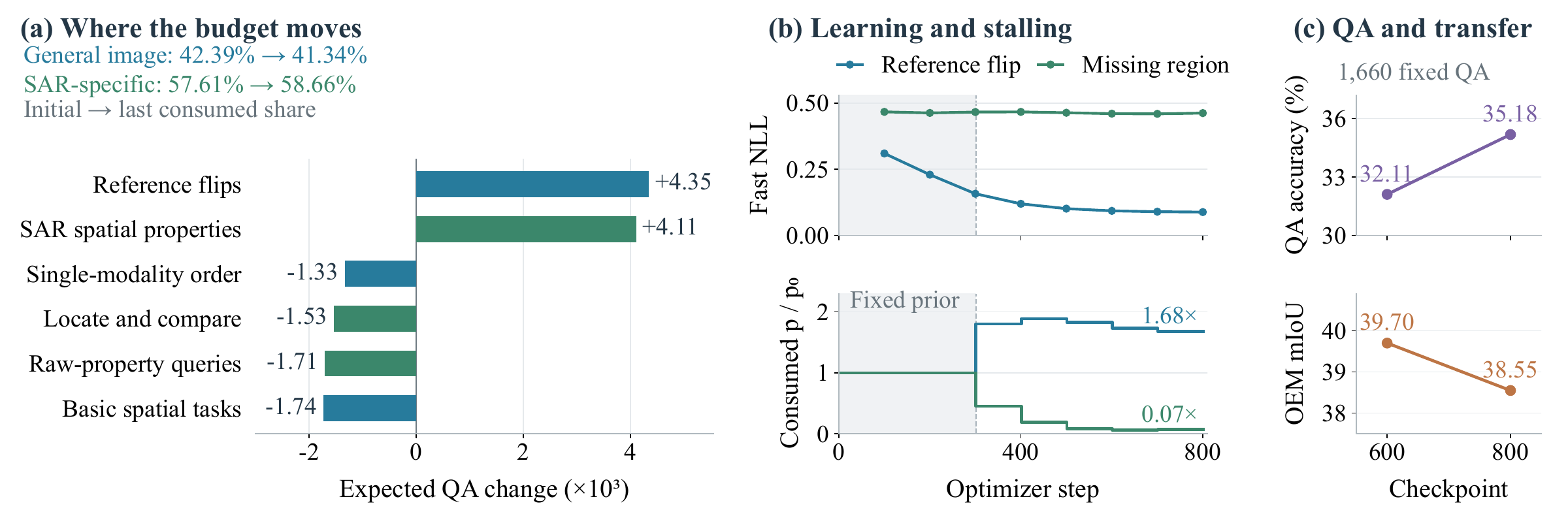}
\caption{\textbf{How adaptive sampling changes training.} (a) Cumulative changes in task exposure, colored by task family. (b) Loss and sampling trajectories for two SARLO tasks. Fast NLL is the fast exponential moving average of answer-token negative log-likelihood. Consumed $p/p_0$ is the ratio of the sampling probability used in training to its initial value. (c) Question-answering (QA) accuracy and OEM-SLC-Amplitude mIoU can diverge.}
\label{fig:sampling_analysis}
\end{figure*}

Table~\ref{tab:sampling_comparison} first examines what supervision makes brief SAR adaptation effective. Both task families improve the original encoder, with SAR-specific tasks contributing more than general image tasks (+1.65 vs.\ +1.04 mIoU). Domain-specific supervision is therefore particularly valuable, yet combining the two families performs better than either alone. The implication for task design is to complement general spatial learning with domain-specific evidence, rather than replace it. Together, these tasks yield a 1.92 mIoU gain under static sampling, showing that computable supervision and short post-pretraining already provide an effective route to specializing an inherited visual representation.

With this complementary task pool established, we next ask how to use it more effectively within the same 600-step budget. Our loss-driven sampler adds 1.49 mIoU and outperforms all compared curriculum strategies, including ACL by 0.83. TSCL and DWA remain below static sampling, indicating that dynamic allocation alone does not guarantee better transfer. The results support two distinct contributions: task construction supplies effective domain supervision, while allocation based on learning state increases its value within a fixed adaptation budget.

\begin{table}[!t]
\centering
\small
\setlength{\tabcolsep}{8pt}
\begin{tabular}{@{}llc@{}}
\toprule
Sampling & Task families & mIoU $\uparrow$ \\
\midrule
Base encoder & \textemdash & 36.29 \\
\midrule
\multirow{3}{*}{Static} & General image only & 37.33 \\
 & SAR-specific only & 37.94 \\
 & Both & 38.21 \\
\midrule
TSCL~\cite{matiisen2017tscl} & \multirow{4}{*}{Both} & 37.75 \\
DWA~\cite{liu2019dwa} & & 37.33 \\
ACL~\cite{graves2017acl} & & 38.87 \\
\textbf{\method }& & \textbf{39.70} \\
\bottomrule
\end{tabular}
\caption{Task-family ablations and sampling strategies on OEM-SLC-Amplitude. }
\label{tab:sampling_comparison}
\end{table}

\subsection{How adaptive sampling changes training}
\label{sec:sampling_analysis}

\begin{table}[!t]
\centering
\small
\setlength{\tabcolsep}{4pt}
\begin{tabular}{cccc}
\toprule
Update & QA draws & Last-used TV & Budget shifted \\
\midrule
400 & 51,200 & 21.85\% & ~5.46\% \\
600 & 76,800 & 22.97\% & 10.55\% \\
800 & 102,400 & 24.53\% & 13.04\% \\
\bottomrule
\end{tabular}
\caption{Reallocation relative to the initial mixture. Last-used TV is the total variation distance $\frac12\sum_g|p_g-p_g^0|$, where $p_g$ is the last-used probability of task variant $g$ and $p_g^0$ its initial probability. Budget shifted measures the fraction of cumulative expected draws reassigned.}
\label{tab:sampling_allocation}
\end{table}

To explain the transfer gain under a fixed budget, we track which questions receive training. The sampler reallocates 10.55\% of cumulative expected draws while changing the SAR-specific share by only 1.05 percentage points (Table~\ref{tab:sampling_allocation}). It therefore refines exposure within the complementary task families, beyond what a single general/SAR mixing ratio can express. Figure~\ref{fig:sampling_analysis}a--b shows the direction of this change: improving reference-flip tasks gain exposure, while stalled cross-modal missing-region tasks receive fewer draws. This concentrates the available updates on questions the model is still learning from, making better use of the same supervision pool.

The resulting loss trends guide allocation, but downstream transfer determines how long adaptation remains useful. From step 600 to 800, diagnostic question-answering (QA) accuracy increases while OEM-SLC-Amplitude mIoU falls from 39.70 to 38.55 (Fig.~\ref{fig:sampling_analysis}c). Continuing to improve visual question answering prediction will not further improve the exported backbone.
\begin{samepage}
\section{Limitations and Future Work}
\label{sec:limitations}

\method opens a path toward rapidly building specialized visual foundation models in domains where annotations are scarce. The key is to replace the SAR-specific questions with computable tasks that capture the target domain's properties, while reusing the general image tasks, MLLM initialization, and efficient adaptation procedure. This offers a route to strong domain encoders without annotated instruction data. Transfer beyond SAR remains untested and requires reliable, visually answerable tasks. Future work will evaluate this recipe across domains and successive MLLM generations under fixed adaptation budgets.

\end{samepage}

\section{Conclusion}
\label{sec:conclusion}

\method reframes SAR foundation modeling as the specialization of an evolving general representation. Beyond an individual backbone, it provides a reusable way to express domain knowledge as computable supervision and adapt the visual capabilities of open MLLMs with modest additional training. This shifts the emphasis of domain-model development toward the supervision needed to make general features useful for SAR, offering a practical route for carrying advances in MLLMs into specialized visual perception.
\label{sec:main_end}

{\small
\bibliographystyle{ieeenat_fullname}
\bibliography{main}
}
\clearpage

\appendix
\setcounter{figure}{0}
\setcounter{table}{0}
\setcounter{equation}{0}
\renewcommand{\thefigure}{S\arabic{figure}}
\renewcommand{\thetable}{S\arabic{table}}
\renewcommand{\theequation}{S\arabic{equation}}
\renewcommand{\theHfigure}{supp.\arabic{figure}}
\renewcommand{\theHtable}{supp.\arabic{table}}
\renewcommand{\theHequation}{supp.\arabic{equation}}
\section{Pretraining Data and Reproducibility}
\label{sec:supp_data}

We use TerraMesh Sentinel-1 GRD~\cite{blumenstiel2025terramesh}, paired SARLO-80 observations~\cite{debuysere2026sarlo}, and unlabeled KuroSiwo GRD/SLC data~\cite{bountos2024kurosiwo}. Table~\ref{tab:supp_sources} lists the training and evaluation source pools. Each record can contain multiple polarizations, dates, or paired views and can generate several self-supervised questions.

\begin{table}[t]
\centering
\small
\setlength{\tabcolsep}{8pt}
\begin{tabular}{@{}lrr@{}}
\toprule
Source & Train & Eval \\
\midrule
TerraMesh & 703,460 & 257,276 \\
SARLO-80 & 101,699 & 8,700 \\
KuroSiwo & 53,889 & 51,152 \\
\midrule
Total & 859,048 & 317,128 \\
\bottomrule
\end{tabular}
\caption{Source records used for SAR adaptation post-pretraining and the evaluation pool from which diagnostic questions are sampled.}
\label{tab:supp_sources}
\end{table}

\subsection{Visible images and measurement targets}

Each record provides native SAR values for computing targets and rendered images for the MLLM. Keeping these representations separate preserves measurement information while supporting a common visual input format.

\paragraph{TerraMesh and KuroSiwo.}
TerraMesh provides $264\!\times\!264$ VV/VH observations in dB on a 10\,m grid. KuroSiwo provides $224\!\times\!224$ chips across three dates, with GRD in linear power and SLC retaining amplitude and phase. Default displays use VV, converted to dB, scaled between the first and 99th percentiles, and resized to $448\!\times\!448$ with three repeated grayscale channels. Measurement tasks use a shared scale that preserves the target relation across views.

\paragraph{SARLO-80.}
We use the official $1024\!\times\!1024$ SAR and optical PNGs together with native complex SAR arrays. Each draw samples a common 256-, 512-, or 1024-pixel window with equal probability and resizes visible inputs to $448\!\times\!448$. Native SAR values determine relative-power targets. Displayed reference pixels determine brightness and contrast targets.  

\begin{table*}[t]
\centering
\small
\setlength{\tabcolsep}{4pt}
\renewcommand{\arraystretch}{1.13}
\begin{tabularx}{\textwidth}{@{}p{0.20\textwidth}Xrrrr@{}}
\toprule
Task group & Task and answer & Var. & T & K & S \\
\midrule
\multicolumn{6}{l}{\emph{General image tasks}} \\
Basic spatial tasks &
Recover SAR patch order, match crops, identify flips or missing patches, and locate polygonal crops. Return a permutation, candidate index, flip descriptor, or box. & 6 & 6 & 6 & 6 \\
Single-modality ordering &
Reorder SAR or optical patches using a same-modality reference, one fixed anchor, or no reference. Return the permutation restoring a $2\times2$ or $3\times3$ grid. & 12 & 6 & 6 & 12 \\
Reference flips &
Compare a SAR image with its unchanged reference to identify the flipped cell in a $2\times2$ or $3\times3$ grid. Return its index and flip direction. & 2 & 2 & 2 & 2 \\
Multiscale matching &
Match a SAR or optical query resized by $0.5\times$ or $1.5\times$ to three same-modality candidate regions. Return the matching index. & 4 & 2 & 2 & 4 \\
\cmidrule(lr){1-6}
Subtotal & & 24 & 16 & 16 & 24 \\
\midrule
\multicolumn{6}{l}{\emph{SAR-specific tasks}} \\
SAR spatial properties &
Compare native-power texture, scattering concentration, edges, clusters, direction, shape, or extent. Return the relevant quadrant, orientation, or shape label. & 7 & 7 & 6 & 6 \\
SAR multiview relations &
Compare VV/VH or temporal SAR views for power ratios, correspondence, texture, change, or stability. Return a candidate, band, quadrant, or change direction. & 5 & 5 & 0 & 0 \\
Cross-modal ordering &
Reorder patches using an opposite-modality reference or a grid mixing SAR and optical patches. Return the permutation restoring the $2\times2$ or $3\times3$ layout. & 12 & 0 & 0 & 12 \\
Paired local tasks &
Use paired SAR--optical views to match crops, select missing regions, detect flipped patches, or locate regions. Return indices, flip descriptors, or boxes. & 16 & 0 & 0 & 16 \\
Pair retrieval and shift &
Retrieve a crop's paired source among geographically separated opposite-modality candidates, or undo a synthetic offset. Return a candidate index or translation vector. & 8 & 0 & 0 & 8 \\
Raw-property queries &
Use a quadrant marked in the optical image to query its paired SAR view. Rank its native power, heterogeneity, or scattering concentration from 1 to 4. & 3 & 0 & 0 & 3 \\
Cross-modal polygons &
Locate a polygon-masked query in an opposite-modality reference. Return its tight axis-aligned box on the normalized $0$--$1000$ coordinate grid. & 2 & 0 & 0 & 2 \\
Mixed candidate matching &
Match one query to four candidates: two SAR and two optical. Return the candidate covering the same region, in either modality. & 4 & 0 & 0 & 4 \\
Interleaved matching &
Match four interleaved SAR/optical queries one-to-one with opposite-modality candidates covering the same regions. Return candidate indices in query order. & 2 & 0 & 0 & 2 \\
Locate and compare &
Locate an unmarked cross-modal query in the reference, then rank its entire containing quadrant by displayed brightness or contrast. Return a rank from 1 to 4. & 4 & 0 & 0 & 4 \\
\cmidrule(lr){1-6}
Subtotal & & 63 & 12 & 6 & 57 \\
\midrule
Total & & 87 & 28 & 22 & 81 \\
\bottomrule
\end{tabularx}
\caption{Tasks used in the main post pretraining, with their inputs, required operation, and answer form. Var. counts distinct configured variants; T, K, and S count active variants for TerraMesh, KuroSiwo, and SARLO. The 14 task groups contain 87 variants and 131 source--task combinations. The two families match the taxonomy used in the sampling analysis.}
\label{tab:supp_task_coverage}
\end{table*}

\subsection{Two task families and their coverage}
\label{sec:supp_tasks}

Table~\ref{tab:supp_task_coverage} describes the 14 task groups. General image tasks comprise 24 variants and 56 source--task combinations; SAR-specific tasks comprise 63 variants and 75 combinations. Sampling operates on individual source--task variants.

Native measurement tasks convert each SAR representation to relative power:

\begin{equation}
 P(u)=\begin{cases}
 |s(u)|^2,&\text{complex SAR},\\
 A(u)^2,&\text{amplitude},\\
 10^{d(u)/10},&\text{dB values},\\
 I(u),&\text{linear intensity}.
 \end{cases}
 \label{eq:power}
\end{equation}
Here $u$ indexes a pixel, and the source metadata determines the conversion. Measurement tasks require native and rendered views to yield the same target. Multiview tasks use TerraMesh. SARLO supplies cross-modal tasks in both directions.

\paragraph{Representative measurement targets.}
For a region $R$ with $n$ native-power pixels and positive mean, mean power, heterogeneity, and scattering concentration are
\begin{align}
 \mu_R &= \frac{1}{n}\sum_{u\in R}P(u),\\
 H_R &= \frac{\frac{1}{n}\sum_{u\in R}(P(u)-\mu_R)^2}{\mu_R^2},\\
 C_R &= \frac{\sum_{u\in T_R}P(u)}{\sum_{u\in R}P(u)},
\end{align}
where $T_R$ contains the $\lceil0.01n\rceil$ highest-power pixels. For heterogeneity and concentration quadrant-selection questions, the winning score must exceed the runner-up by more than 5\% of the winner's magnitude. Raw-property ranking questions require adjacent sorted scores to differ by at least 3\% of the largest score magnitude. Questions are rejected when native and displayed views disagree on the answer.

\begin{figure*}[t]
\centering
\includegraphics[width=\textwidth]{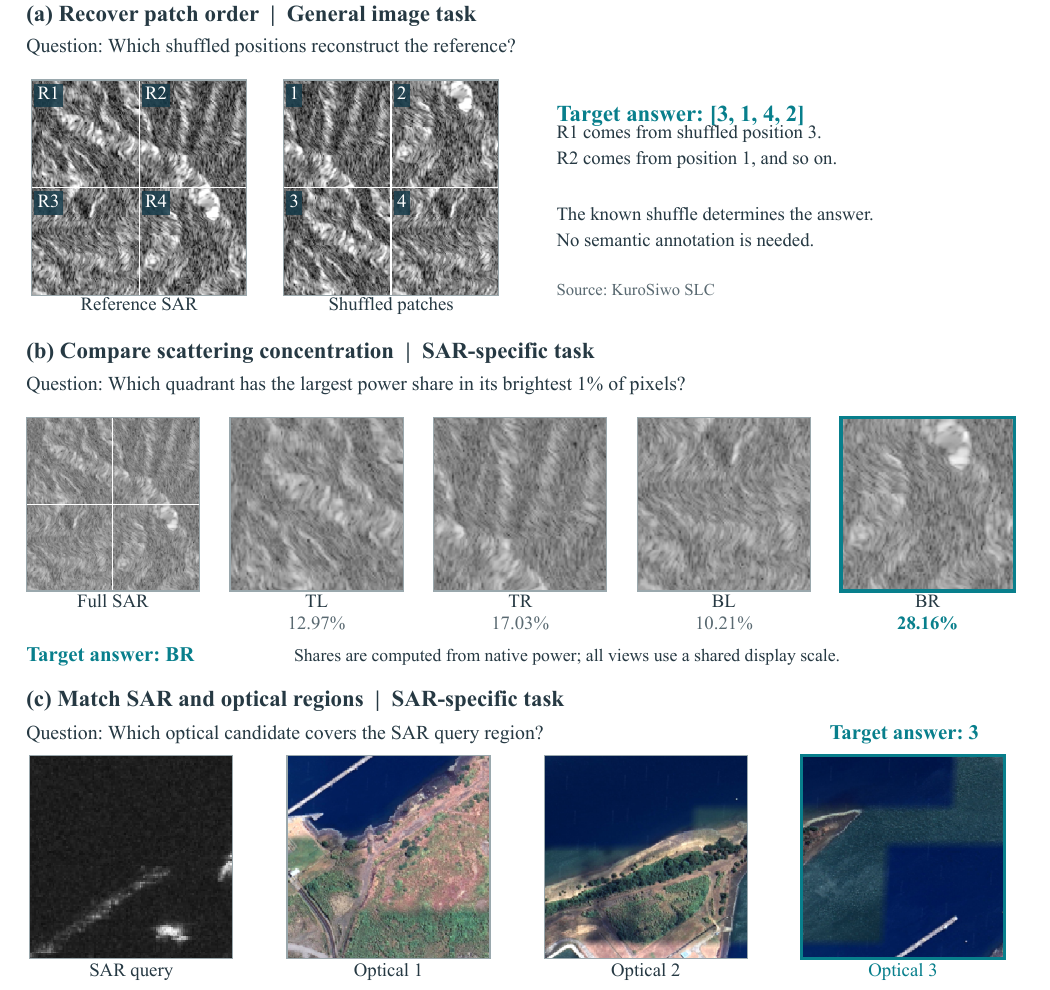}
\caption{Representative self-supervised questions from real KuroSiwo and SARLO observations. (a) The known patch permutation determines the answer. (b) The fraction of native power carried by each quadrant's brightest 1\% of pixels determines its scattering concentration. TL, TR, BL, and BR denote the top-left, top-right, bottom-left, and bottom-right quadrants. (c) Paired crop coordinates identify the optical region containing the SAR query. Grids, labels, and highlighted answers are explanatory overlays.}
\label{fig:supp_task_examples}
\end{figure*}

Figure~\ref{fig:supp_task_examples} illustrates three ways to obtain supervision: invert a known image operation, compute a native SAR statistic, or use paired crop coordinates. Together they show how the task pool combines general spatial reasoning with SAR-specific measurements and cross-modal relations.

\section{Adaptation Settings}
\label{sec:supp_training}

Table~\ref{tab:supp_training} gives the 600-step SWIFT~\cite{zhao2025swift} adaptation recipe used to produce the final visual backbone.

\begin{table}[t]
\centering
\small
\setlength{\tabcolsep}{3pt}
\begin{tabularx}{\linewidth}{@{}p{0.50\linewidth}X@{}}
\toprule
Setting & Main experiment \\
\midrule
MLLM & Qwen3.5-2B~\cite{qwen2026qwen35} \\
LoRA rank / $\alpha$ / dropout & 16 / 32 / 0 \\
Adapter placement & All linear layers \\
Learning rate: vision / merger & $5\!\times\!10^{-5}$ / $5\!\times\!10^{-5}$ \\
Learning rate: language & $1\!\times\!10^{-5}$ \\
Schedule / warmup & Constant / none \\
Optimizer & AdamW~\cite{loshchilov2019adamw} \\
Adam $\beta_1,\beta_2$ / $\epsilon$ & $(0.9,0.95)$ / $10^{-8}$ \\
Weight decay / gradient clipping & 0.01 / 1.0 \\
Global batch / updates & 128 / 600 \\
Run parallelism & 2 GPUs, batch 2 per GPU \\
Gradient accumulation & 32 \\
Model / adapter precision & bfloat16 / float32 \\
Maximum sequence length & 8192 tokens \\
Random seed & 3407 \\
\bottomrule
\end{tabularx}
\caption{Main ViT-L adaptation configuration.}
\label{tab:supp_training}
\end{table}

\subsection{Downstream optimization}

The exported ViT-L has 24 layers and width 1024; ViT-B has 12 layers and width 768. OEM comparisons use batch size 8 and AdamW, with learning rates of $2\times10^{-4}$ for the head and $10^{-6}$ for the visual layers. All OEM comparisons report the final fine-tuning checkpoint at epoch 50.

WHU-OPT-SAR follows the SARATR-X-v2 scene partition~\cite{li2026saratrxv2}: scenes are shuffled with seed 42 and divided 60/20/20 into training, validation, and test sets before tiling. Detection uses batch size 8, weight decay 0.05, and layer decay 0.95. Learning rates are $2.5\times10^{-5}$ for RSAR/M4-SAR and $10^{-5}$ for SARDet-100K, reduced at epochs 8 and 11. For SARDet-100K, newly initialized adapter, neck, RPN, and RoI-head parameters use a $3\times$ learning-rate multiplier. M4-SAR uses anchor scales $\{1,2,4,8\}$ to cover small objects. Segmentation uses batch size 16, a $10^{-4}$ learning rate with a visual-core multiplier of 0.1, and polynomial decay. It uses random training crops and sliding-window inference. 

\section{Task Sampling Details}
\label{sec:supp_sampling}

\subsection{From answer losses to task probabilities}

The prior $p_g^0$ first selects TerraMesh, SARLO, or KuroSiwo with weights $1{:}4{:}1$, then an available implementation group uniformly, and finally a variant using its normalized within-group weight. These internal groups differ from the two task families in Table~\ref{tab:supp_task_coverage}. Each group receives the same total prior within its source, regardless of its number of variants.

We aggregate detached mean answer-token NLL by source--task variant. For a fixed variant, let $t$ index statistical updates, each using at least eight examples. With mean $\mu_t$ and variance $v_t$ over $n_t$ examples, the noise scale is $e_t=\sqrt{v_t/\max(1,n_t-1)}$. Omitting the variant subscript $g$, we update fast and slow averages as
\begin{align}
 F_t&=(1-\alpha_f)F_{t-1}+\alpha_f\mu_t,\\
 S_t&=(1-\alpha_s)S_{t-1}+\alpha_s\mu_t,
\end{align}
with $\alpha_f=0.6$ and $\alpha_s=0.1$. Both averages start at the first valid mean, which also sets the fixed reference $R=\max(\mu_1,\epsilon)$ with $\epsilon=10^{-6}$. Residual loss and progress are
\begin{align}
 r_t&=\clip_{[0,1]}(F_t/R),\\
 q_t&=\clip_{[-1,1]}\!\left(
 \frac{\sgn(S_t-F_t)[|S_t-F_t|-0.8e_t]_+}
 {\max(S_t,\epsilon)}\right).
\end{align}
The clipping limits $r_t$ to $[0,1]$ and $q_t$ to $[-1,1]$.

Equation~\ref{eq:utility} uses coefficients $1$, $3$, and $3$ for the residual, progress, and rebound terms. The exploration mixture in Eq.~\ref{eq:sampling} uses $\eta=0.01$. We smooth the target distribution with the current probabilities as $z_g=0.4p_g^{\rm old}+0.6d_g$. If a source has zero weighted utility, its target returns to the base distribution. The update is constrained by
\begin{align}
 l_g&=\max(0.01p_g^0,p_g^{\rm old}/5),\\
 h_g&=\min(12p_g^0,5p_g^{\rm old}),\\
 p_g^{\rm new}&=\clip_{[l_g,h_g]}(a_s z_g),
 \qquad \sum_{g\in\mathcal G_s}p_g^{\rm new}=M_s.
\end{align}
A source-specific multiplier $a_s$ enforces its fixed total probability. The bounds maintain exposure and limit changes relative to the preceding distribution and initial prior.

Training proceeds in windows of 100 optimizer updates, each containing 12,800 question draws. Every source--task variant must have at least two valid statistical updates before its source is reallocated. Feedback at the end of window $k$ determines window $k+2$, since the intervening window is already prepared for data loading. Training minimizes the answer loss without importance reweighting. Sampling changes task exposure. 

\subsection{Analysis protocol}

For the training-dynamics analysis only, we extend the adaptation run to 800 updates.

Figure~\ref{fig:sampling_analysis} and Table~\ref{tab:sampling_allocation} use distributions actually consumed during training. The cumulative shifted budget is
\[
 \frac{1}{2N}\sum_g\left|\sum_w N_wp_{g,w}-Np_g^0\right|,
 \qquad N=\sum_wN_w,
\]
where $N_w$ is the number of draws in window $w$. The diagnostic panel contains 1,660 questions from the source--task pairs used in training. Its reported score is the unweighted mean of the accuracies of the 87 active task variants. Most variants require an exact match to the structured answer, while some box-localization variants use an IoU threshold of 0.5.

\subsection{Implementation of comparison strategies}

The strategies compared in Table~\ref{tab:sampling_comparison} operate within the same fixed source budgets. Teacher--Student Curriculum Learning (TSCL)~\cite{matiisen2017tscl} uses the magnitude of a smoothed recent learning-progress estimate. Dynamic Weight Average (DWA)~\cite{liu2019dwa} converts consecutive-loss ratios into sampling probabilities, and Automated Curriculum Learning (ACL)~\cite{graves2017acl} uses an Exp3.S curriculum. DWA was originally proposed for loss weighting. Here its weights control task sampling. All strategies share the main run's data and optimization settings and are evaluated after 600 adaptation updates.

TSCL uses a temperature of 0.01, a five-observation history, and an EMA coefficient of 0.5. DWA uses a temperature of 2. ACL uses a learning rate of 0.1 and a reward reservoir of 2,048 observations.

\paragraph{Task-family ablations.}
General-only and SAR-specific-only retain the corresponding family, preserve the TerraMesh:SARLO:KuroSiwo source weights of $1{:}4{:}1$, and renormalize the retained task priors within each source. Both uses the complete task pool. All three conditions use static sampling and share the initialization, rank-16 adapters, optimizer, global batch size of 128, and 600-update budget.

\paragraph{Sampler parameter selection.}
Preliminary tuning of our dynamic sampler favored moderate reallocation: both more conservative and more aggressive settings yielded weaker downstream transfer in these trials. We therefore use the reported configuration to balance responsiveness to learning progress with stable task exposure.

\section{A Compact Dataset for Rapid Downstream Validation}
\label{sec:supp_oem}

OEM-SLC-Amplitude pairs native Umbra SLC amplitudes with OpenEarthMap-SAR labels projected onto the slant-range grid~\cite{xia2025openearthmapsar}. It has seven classes and $512\!\times\!512$ chips, with three channels encoding amplitude in dB and spatially averaged power at two scales.
The development set contains 274 training and 654 evaluation chips. It also shares labels and regions with OpenEarthMap-SAR. The OEM-SLC-Amplitude will be released after the paper is accepted.

\end{document}